# Anatomy and Physiology of Artificial Intelligence in PET Imaging

Tyler J. Bradshaw, PhD, and Alan B. McMillan, PhD

*Abstract*— The influence of artificial intelligence (AI) within the field of nuclear medicine has been rapidly growing. Many researchers and clinicians are seeking to apply AI within PET, and clinicians will soon find themselves engaging with AI-based applications all along the chain of molecular imaging, from image reconstruction to enhanced reporting. This expanding presence of AI in PET imaging will result in greater demand for educational resources for those unfamiliar with AI. The objective of this article to is provide an illustrated guide to the core principles of modern AI, with specific focus on aspects that are most likely to be encountered in PET imaging. We describe convolutional neural networks, algorithm training, and explain the components of the commonly used U-Net for segmentation and image synthesis.

## I. INTRODUCTION

Artificial intelligence (AI) has seen an explosion in interest within the field of nuclear medicine [1]. This interest has been driven by the rapid progress and eye-catching achievements of machine learning (ML) algorithms over the past decade. AI, and in particular computer vision, is now receiving attention from many outside of the field of computer science, hoping to apply the promising technologies within their own field of study. Nuclear medicine, like many other medical specialties, is poised to benefit from AI in a number of ways [2-4]. But newcomers to ML may be overwhelmed by the nearly limitless acronyms, network architectures, and publications claiming "state-of-the-art performance", all of which can be challenging for beginners to navigate.

The goal of this article is provide an illustrated guide to foundational concepts in AI. Given the breadth of AI, this manuscript will focus on topics and networks that are most relevant to PET imaging. There are many classes of AI algorithms, many of which are beyond the scope of this article. For example, there are supervised learning algorithms that are trained using datasets with paired inputs and labels, and unsupervised learning algorithms that learn relationships using unlabeled data (e.g., clustering). Algorithms can also be classified based on application, such as computer vision algorithms that are applied to images, and natural language processing algorithms that are applied to text. Algorithms can be further classified according to the structure and function of the network, such as artificial neural networks, decision forests, support vector machines, transformers, and so on. Algorithms are designed with certain structures (anatomy) and functions (physiology) so that they are capable of handling specific tasks (e.g., image classification). This article will focus on supervised learning algorithms that process images, with a specific focus on convolutional neural network, as these are currently the most relevant algorithms to PET imaging. The target audience is non-technical, future consumers of AI algorithms in nuclear medicine who wish to better understand this emerging technology. We aim to convey a high level conceptual understanding of AI, whereas readers interested in a deeper treatment of its mathematical underpinnings are referred to other publications [5-7].

The article is organized as follows. First, we provide an overview of the pipeline for AI algorithm development. We then give a step-by-step survey of the components and operations of AI algorithms, particularly the basic convolutional neural network (CNN), which is the principal ingredient of most modern computer vision algorithms. We then describe the process of model training. Finally, we explain the components of the widely used U-Net architecture, which is arguably the most widely used CNN in the medical imaging community [8].

## II. STEPS OF ALGORITHM DEVELOPMENT

There is a common pipeline used when developing ML algorithms. Understanding the development life cycle of an ML algorithm is important for placing promising studies or newly FDA-cleared AI software in proper context. The steps of the pipeline begin once an investigator has a clearly defined task that they would like to perform (which can itself be a challenge). The task is a prediction task: given some input data, the model predicts the expected output. To build the prediction model, investigators will then collect data, label data, build the network, train the model, evaluate the model, and deploy the model. This pipeline is shown in Figure 1.

Each step of the pipeline is deserving of its own in-depth treatment. Indeed, many review and educational articles have been dedicated to the individual steps [9-11]. When developers take shortcuts or fail to follow best practices anywhere along the pipeline, they risk seeing their algorithms fail during evaluation or post deployment [12]. Most of the important terms and concepts of ML algorithm development are beyond the scope of this article, however many of them are listed in Figure 1 and readers are encouraged to explore them further. Also, a glossary of terms used in this article is found in Table 1

Tyler J Bradshaw and Alan B McMillan are from the University of Wisconsin, Department of Radiology, Madison, WI, USA (e-mail: tbradshaw@wisc.edu). TB and AM receive research support from GE Healthcare.

Table 1. Glossary of Terms

| Term | Definition |
| --- | --- |
| 1×1 convolution | A convenient tool for changing the number of channels in a layer |
| Activation function | A function that transforms the output of a layer; often used to add non-linearity to a network |
| Activation (feature) map | The result of network operations performed on input data; often represent salient features of the input data |
| Backpropagation | A method used by an optimizer to compute the network's gradients |
| Batch (mini-batch) | Datasets are partitioned into batches for training; one batch is used during each iteration |
| Batch normalization | An operation that normalizes the values of an activation map |
| Channel | The number of activation maps or input depth; for images, the size of the 3rd dimension (e.g., red-green-blue = 3 channels) |
| Concatenation | Often used to stack two sets of activation maps together |
| Convolution layer | A network layer in which a bank of filters is convolved with an input matrix, producing an activation map for each filter |
| Cross validation | A data partitioning technique where the dataset is repeatedly sampled into different training and testing sets |
| Decision tree | A series of binary yes/no operations performed on input data; trees are often combined together to create decision forests |
| Encoder-decoder | A type of network that consists of a series of downsampling operations followed by a series of upsampling operations |
| Epoch | Passing over the entire dataset (all batches) during training |
| Filter | A set of weights that is convolved with an input image as part of a convolution layer; updated during training |
| Fully-connected (dense) layer | A single layer of a traditional neural network; each node connects to each element of the input |
| Generative adversarial network | A framework for training models by having two models compete and learn from each other |
| Hyperparameters | Model parameters not explicitly learned during training; often design choices (e.g., number of layers, channels, epochs) |
| Iteration | An individual update step during training, often using a single batch of data |
| Linear layer | An activation function that only scales the input; used for continuously valued outputs |
| Loss function | A measurement of how far off the model's predictions are from the labels. Used to guide model training. |
| Max pooling | A downsampling operation that passes on the highest value in each patch after an image has been partitioned into patches |
| Neuron/node | A single operation within an artificial neural network |
| Optimizer | The method used to update the model's weights based on the loss function |
| Overfitting | The network memorizing training examples; often results in poor generalization performance |
| Rectified linear unit (ReLU) | An activation function that sets all negative valued inputs to zero and passes through all positive values |
| Sigmoid | An activation function that yields a value between 0 and 1; used for binary classification |
| Softmax | An activation function that provides the probabilities for each possible class the sample might belong to; for multi-class classification |
| Stochastic gradient descent | A commonly used optimizer |
| Supervised learning | Methods by which an algorithm learns to map input data to a desired output by training with a dataset of paired input-label examples |
| Tensor | A data object; often a multidimensional data array |
| U-Net | An encoder-decoder network that is commonly used for segmentation and image synthesis |
| Upsampling (up-convolution, transpose convolution) | An operation that increases the dimensions of the input data |
| Unsupervised learning | Methods by which an algorithm learns patterns in an unlabeled dataset (e.g., clustering) |
| Weights (parameters) | Coefficients that are used in network operations and are updated/learned during training |

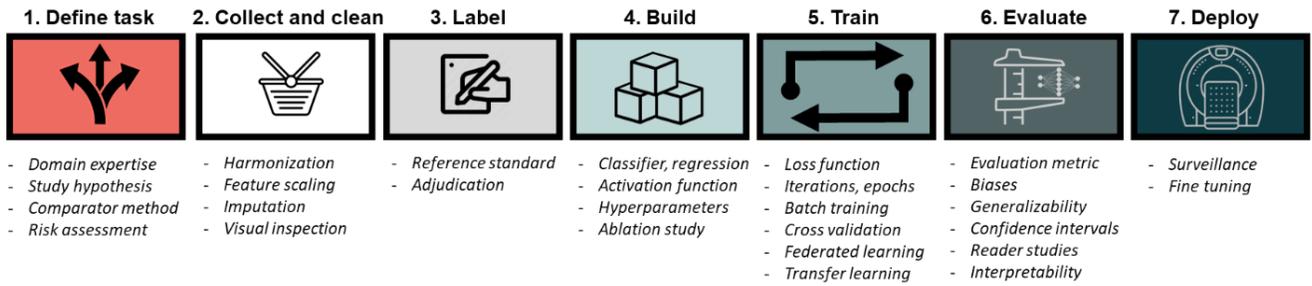

Fig. 1. The steps of machine learning algorithm development, together with key concepts for each step. Many of the key concepts are beyond the scope of this article and readers are encouraged to explore these concepts further through other sources.

Data collection and labeling are arguably the most critical but time-consuming steps of building a ML model. If the dataset does not reflect the clinical task (e.g., using radiotherapy contours for PET lesion detection algorithm) or the clinical patient population (e.g., lacking obese patients), then the algorithm will likely reflect those limitations. High quality, large, and diverse datasets are needed for ML algorithm development.

A key concept for users to understand is *generalizability*, together with its counterpart *overfitting*. The primary goal -- and also the primary challenge -- of ML algorithm development is to create an algorithm that performs well when applied to unseen data (i.e., data not available to the model during training). ML algorithms can easily memorize training samples: they can detect noise patterns or features that are highly specific to the training dataset and then rely on those features to make predictions. But those features will not be useful, and in fact will be misleading, when used to make predictions for a new dataset. Therefore, significant effort is spent during algorithm development to preventing overfitting. Data also needs to be collected and labeled from diverse sources, matching the diversity of the expected population, to avoid overfitting to a specific subpopulation. Then, once the model is trained, its performance must be evaluated with new data that is external to the development dataset [13].

### III. BUILDING BLOCKS OF MACHINE LEARNING NETWORKS

The primary assumption underpinning many ML algorithms is that simple mathematical operations, when intelligently stacked together, can be used to represent highly complex relationships between input data and training labels. Motivated by the "simple" functions of individual biological neurons in the brain, many modern ML networks are in fact very long series of simple operations. The operations rely on numerical weights or parameters that are learned during training. These building blocks often include weighted sums, or binary yes/no decisions, or convolutions, as illustrated in Figure 2. Most modern ML networks are primarily composed of these simple operations, such as artificial neural networks (weighted sums), random forests (binary decisions), and CNNs (convolutions). The building blocks are typically combined with additional operations (e.g., non-linear functions) and organized into complex pathways to better represent the complex relationships they are tasked to learn.

### IV. THE CONVOLUTIONAL NEURAL NETWORK

This section walks the reader step-by-step through all the major components of a CNN.

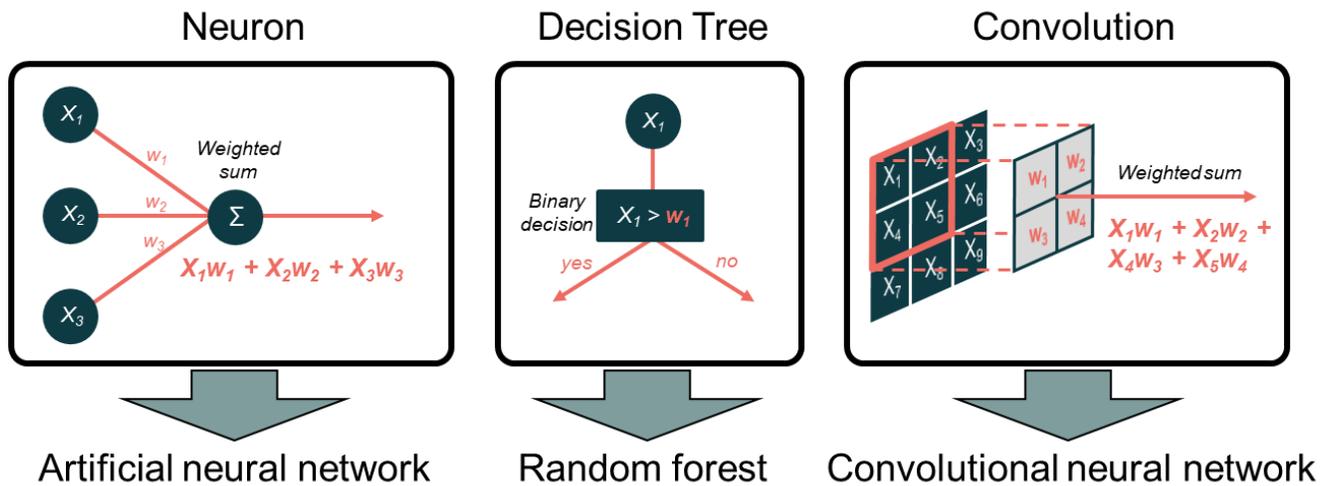

Fig. 2. Relatively simple functions are used as the main building blocks of most AI networks. Learnable weights ($w_i$) are used to perform basic operations on input data ($X_i$). When stacked together in large numbers, these building blocks can create complex and powerful networks.

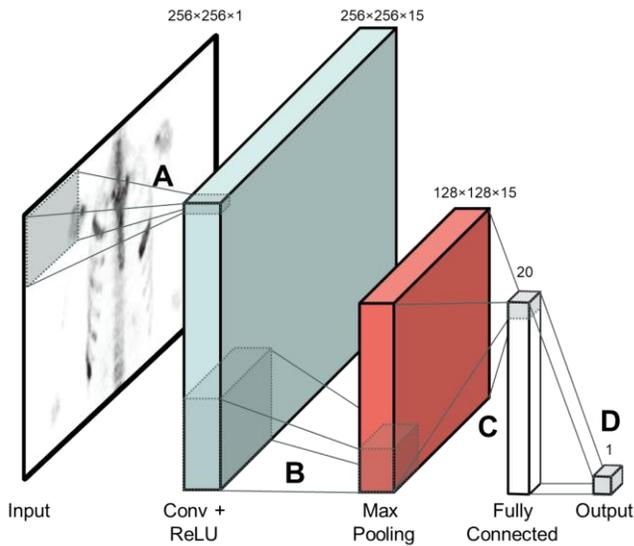

Fig. 3. This example convolutional neural network is dissected throughout the article. The key steps are labeled: *A* is the convolution operation (Figure 4), *B* is max pooling (Figure 5), *C* is a fully connected layer (Figure 6) and *D* is the output (Figure 7). The numbers above each layer indicate the dimensions of the activation maps or number of nodes: Nx×Ny×Nchannels.

*Diagrams*

CNNs are often represented using diagrams like the one shown in Figure 3. There are different conventions for representing networks in literature, which can be a source of confusion. In general, diagrams will either illustrate the series of operations that are performed on the input data (e.g., showing a series of convolution operations as in He *et al* [14]) or diagrams will illustrate the data that result from the network operations (e.g., showing how the dimensions of the data change following network operations as in Simonyan *et al* [15]). The latter style is used in Figure 3. Often the weights of a network, such as the bank of convolutional filters, are not represented in the diagrams but are implied.

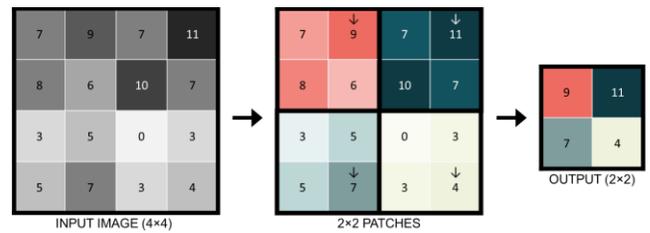

Fig. 5. The max pooling operation partitions an image into patches (in this case 2×2 patches) and then passes along the highest value in the patch to the next layer. This has the effect of downsampling the image/activation map.

*Convolution layers*

The network depicted in Figure 3 is a simple 3-layer CNN. The 3 components of this network that are considered "layers" are the convolution layer, the fully-connected layer, and the output layer. This small network will serve as a toy example that will allow us to understand the foundational components of a CNN. In the figure, each operation performed by the network is represented by letters *A-D*. Each operation is discussed below and illustrated in more detail in Figures 4-7.

The first convolution operation is labeled *A* in Figure 3. This is a 2D convolution and is depicted in greater detail in Figure 4.

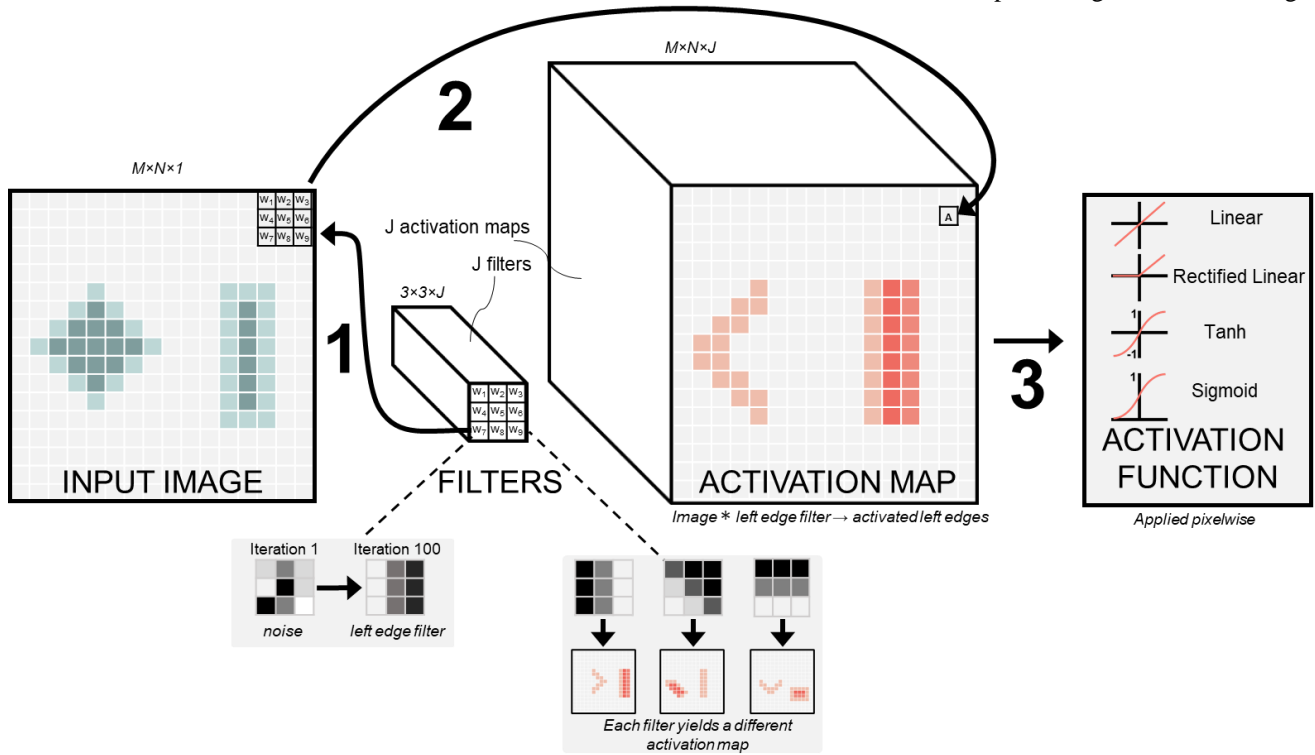

Fig. 4. A convolutional layer consists of 3 operations, labeled 1-3 in the figure. 1) A bank of filters is convolved with the image in a sliding window fashion. The filters start as random numbers and evolve over the course of training to become feature detectors, such as edge detectors. Each of the *J* filters is convolved with the input image. 2) The convolution operation produces an activation map. The activation map reflects the locations of the input image that contain whichever feature the filter has learned to detect (e.g., the left edge is activated by left edge filters). Each filter produces a different activation map. 3) An activation function is applied pixel-wise to the activation maps so that the network is capable of learning non-linear relationships between the input image and the training label.

The first convolution layer takes two arrays as input: the input image and a bank of learnable 2D filters or kernels. The convolution layer produces a third array as output: the activation map. Each filter (the number of filters is a parameter selected by the developer) is convolved with the input image and consequently creates its own independent and unique activation map as output: $J$ filters produce $J$ activation maps. The filters are conceptually understood to represent "feature detectors". They begin as random numbers and then during training they evolve into useful filters, such as edge detectors. The activation maps (also called feature maps or internal representations) reflect the part of the input image that is "activated" by the filter. For example, an edge-detecting filter will produce an activation map with high values in the pixels that correspond to edges in the input image. This is depicted in Figure 4. Activation maps in the early layers of the network are thought to reflect simple features (e.g., edges) while activation maps deep within the network are thought to reflect high-level, abstract features (e.g., ribcage). By convention, the number of activation maps produced in a given layer is also called the number of *channels* in that layer (100 activation maps = 100 channels).

*Activation functions*

Following each convolution layer, the resulting activation map is passed through an activation function. The activation function is a mathematical function applied to each element of the activation map (i.e., pixel-wise). The purpose of the activation function is to introduce non-linearity into the network, otherwise the sequence of convolutions (which are linear operations) would be limited to only learning linear relationships between the input data and target labels. There are a variety of functions that can serve as activation functions, the most common being the rectified linear unit (ReLU) and the sigmoid, as shown in Figure 4. The ReLU is a simple function whose output is 0 if the input (i.e., pixel in the activation map) is less than 0, otherwise its output is identical to the input. The ReLU effectively sets all negative values in the activation maps to zero, which is surprisingly simple yet effective.

Convolution operations are often repeated multiple times in a row. The activation maps resulting from the first convolution layer act as the input for a second convolution layer. These activation maps are convolved with a new, independent bank of learned filters, yielding another set of activation maps in the second layer. Some CNNs have dozens of convolution layers.

*Pooling layers*

Pooling layers, labeled *B* in Figure 3, are a key component of most CNNs. Pooling layers are downsampling operations, which are necessary due to the GPU's limited memory. A succession of convolution layers producing multi-channel activation maps can quickly surpass the GPU's memory capacity when applied to high-resolution images. For a given memory budget, the empirical rule-of-thumb is that one can get greater performance when activation maps are downsampled. This saves memory, which effectively allows one to build a deeper network with more layers and/or channels. Large networks typically have several pooling layers.

The most common type of pooling is max pooling, which is depicted in Figure 5. In 2×2 max pooling, the activation map is first partitioned into 2×2 patches. The highest pixel value in each patch is passed to the next layer while the other 3 pixel values are thrown away. This has the effect of reducing the dimensions of the activation maps by half (512×512 → 256×256) and memory by a factor of 4. Conceptually, by passing the maximum pixel value in a patch to the next layer, the network is indicating to the next layer that there is a high activation (i.e., there is an important feature) within the patch.

*Fully connected layers*

Fully connected (FC) layers are often used prior to the final output of the network, as indicated by label *C* in Figure 3. A FC layer, or dense layer, is equivalent to a traditional artificial neural network comprised of neurons or nodes. The FC layer is depicted in detail in Figure 6. Often the purpose of placing a FC layer at the end of a CNN is to learn complex relationships between the last set of activation maps and the final output of the network. The developer selects the number of nodes in the FC layer, and each element of the input to the layer, which we will call $F_i$, is conceptually "connected" to each node in the FC layer. These connections are often illustrated in diagrams by lines connecting two nodes. Each connection is assigned a learnable weight, $w_i$. The output of each node in the FC layer is a weighted sum, meaning the total sum of all input values multiplied by their respective weights: $\Sigma(F_i \times w_i)$. In practice, this is simply a matrix multiplication of the input vector and the weight vector. Often a learned bias value, $b$, is added to the output of the node and then it is passed through an activation function, as shown in Figure 6. Note that in the figure, the input to the FC layer is not the activation map itself, but rather a flattened version of the activation map, in which the 3-dimensional array gets collapsed to a 1×N vector. This is a necessary step before feeding the activation map to the FC layer.

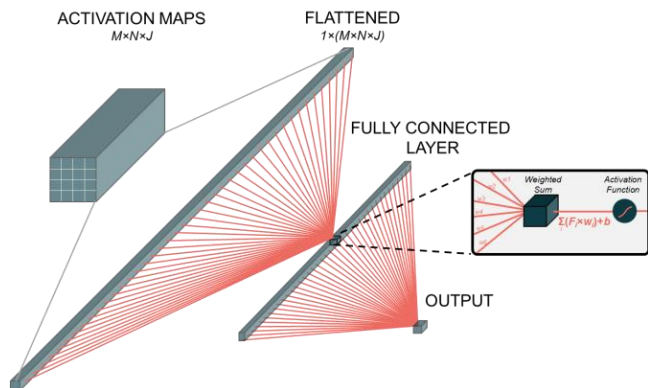

Fig. 6. Fully-connected (FC) layers are often used prior to the output layer of a CNN. The activation maps are first flattened (i.e., converted to a 1-dimensional vector) and then fed to a FC layer. Each element of the flattened activation maps, $F_i$, is connected to each node of the FC layer, with each connection assigned a learnable weight, $w_i$ (for convenience, this figure only shows $F_i$ connected to a single FC node). The output of a node is the weighted sum of all elements feeding into the node, plus a learned bias weight, followed by an activation function.

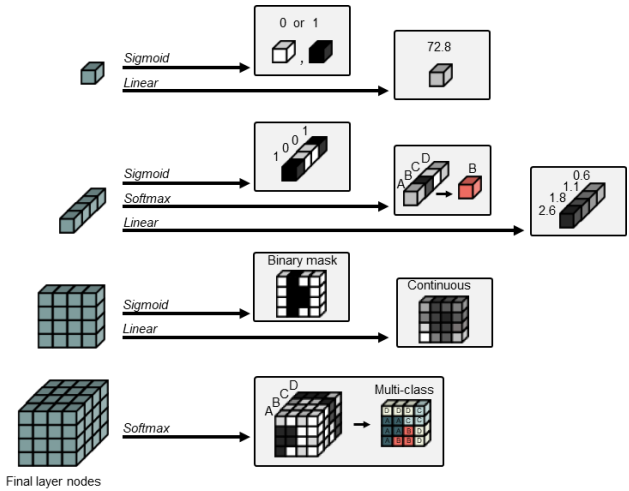

Fig. 7. The output node and its activation function dictate the output of the network. The output layer can have a single node, a vector of nodes, a 2D array of nodes, or (for multi-class segmentation using softmax) a 3D array of nodes.

*Final layer*

The final layer of a network together with its corresponding activation function dictate the output of the model. For our toy network in Figure 3, the final layer is labeled *D* and consists of a single node. The last layer can be a single node, a vector of nodes, or even a 2D or 3D array of nodes, depending on the desired output of the model. The different possible forms that the last layer might take and options for activation functions are depicted in Figure 7. The most common activation functions for the last layer are sigmoid, linear, and softmax. Sigmoid activation functions are ideal for binary tasks (e.g., disease present or disease absent) [7]. The output of a sigmoid is a continuous value between 0 and 1 (see Figure 4), but in practice its output gets rounded to either 0 or 1. Sigmoids are used for binary classification or 2-class image segmentation (for which the last layer would have dimensions $N_x \times N_y$). Linear activation functions are for continuously valued outputs, such as predicting a risk score or for image synthesis. Softmax activation functions are used in multiclass classification problems. Softmax returns a probability score for each of the possible output classes, and the class with the highest score is then used as the model output. Layers using softmax must therefore have an extra dimension whose size is the number of possible classes (see Figure 7). For example, if classifying an image into one of four disease groups, softmax would output four values, one for each disease group, representing the probability of the image belonging to each group. So the final layer should be of dimensions 1×4.

## V. NETWORK TRAINING

A detailed treatment of network training is beyond the scope of this article, but the following is a brief overview of the key components.

*Optimization*

Training is an optimization process. This means that with each iteration of training, the weights of the network (which start off as random numbers) are tweaked in a way that makes the output of the network better (i.e., the network's output gets closer in value to the labels of the training data). For the network shown in Figure 3, the weights that get updated during training are the convolutional filters and the FC layer weights (both of which are not depicted in the figure). Weights can number in the tens of millions for large networks.

An optimization problem requires a loss function. A loss function quantifies the difference between the model's output and the training labels. The overall goal of training is to minimize the network's loss function. For example, a potential loss function for problems with continuously valued outputs might be mean squared error: (*predicted – true*)$^2$. For classification problems, *classification accuracy* could be used as a loss function, but researchers have found that a measure called *cross-entropy* produces better results for classification and is therefore more widely used [7].

The training algorithm, or the optimizer, is the method that determines the magnitude and direction that each weight should be changed during training so that the loss function gets smaller. For example, stochastic gradient descent (SGD) is a commonly used optimizer for CNNs [16]. Most optimizers, including SGD, use the backpropagation algorithm to compute the network's gradients (the gradients tell you how a directional change in a weight will impact the loss function).

*Training considerations*

During training, precautions are taken to prevent the network from overfitting the training dataset. To avoid overfitting, developmental datasets are often partitioned into 2 or 3 distinct sets: the training set for training the model, the test set for estimating the performance of the model on unseen data, and (if needed) a validation set for model selection. The validation set lets developers try out different models without the risk of accidentally selecting a model that is by chance overfitting the test set. An even better method than one-time data splitting is the use of cross validation (CV), in which the dataset is repeatedly split into training and testing sets and a new model is trained with each split. With CV, the whole dataset is used for training and for testing across the different splits.

Due to memory constraints, developers often train the model on batches of examples from the training dataset, updating the model with each batch of data. When enough batches have passed through the model to encompass the full size of the entire training dataset, this is called an epoch. Model training typically proceeds through many epochs and stops when the error in the validation set begins to rise (even if the error in the training set continues to improve).

## VI. THE U-NET

Most CNNs used in academic or commercial applications are more complicated than the toy network shown in Figure 3. In this section, we describe in detail the well-known and widely used U-Net architecture. The U-Net architecture was proposed by Ronneberger *et al*. in 2015 and is depicted in Figure 8 [8]. The U-Net forms a *U* shape, with the left-hand descending side called the encoder, and the right-hand ascending side called the decoder. The U-Net was developed to solve a problem that was challenging at the time: how to produce high-resolution segmentation maps using CNNs. The "funnel" shape of CNNs, in which high resolution images get repeatedly downsampled to low-resolution activation maps, causes the high-resolution information (e.g., sharp edges) to get lost along the way. The U-Net solves this problem in two ways: upsampling and concatenation operations. Figure 8 shows that the upsampling operations begin halfway through the network. An upsampling operation doubles the *X-Y* dimensions of an activation map, and if repeated enough, eventually allows for the network's output (e.g., a segmentation map) to match the dimensions of the input image. It also allows for features from the lowest-resolution level of the network, which represent the most abstract and contextually-aware features, to be passed to the high-resolution output map. Upsampling can be performed using nearest neighbor upsampling, in which 1 pixel value is replicated 4 times into a 2×2 patch, or through up-convolution (also called transpose convolution or deconvolution), in which 1 pixel gets multiplied by 4 learned weights to produce a 2×2 patch. The number of channels often remains the same after upsampling. Note that in Figure 8 the channel depth (i.e., number of activation maps) is not accurately depicted: all layers appear to have the same number of channels. In fact, the U-Net increases in channel number with each stage of the encoder, and then decreases in channel number with each stage of the decoder. The figure ignores this for convenience only.

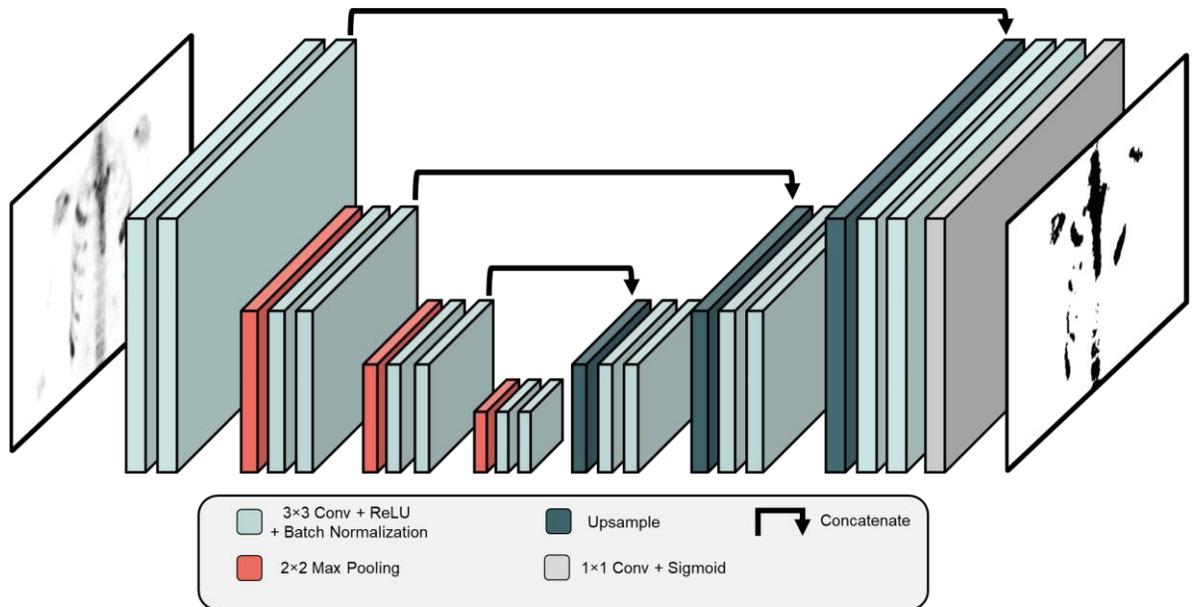

Fig. 8. The U-Net is an encoder-decoder network, with the encoder (left) side consisting of downsampling operations via max pooling and the decoder side consisting of upsampling operations. This U-Net is producing a binary segmentation map from a nuclear medicine image.

The second novel component of the U-Net is the concatenation operations. These operations allow the network to combine the abstract features learned at the lowest resolution layer with the high resolution features of the input images. It works in the following manner: the activation maps on the encoder side of the network, which are produced by convolutional layers acting on the input image and its descendants, are combined with the activation maps on the decoder side of the network, which are produced by upsampling the lowest-resolution features. Because the *X* and *Y* dimensions of some activation maps match on both the encoder and decoder sides of the network, the activation maps can be combined by simply stacking them together channel-wise. In other words, if the activation maps in the encoder side are of dimensions $N_x \times N_y \times N_A$, and the activation maps on the decoder side are $N_x \times N_y \times N_B$, the concatenated maps will be of dimension $N_x \times N_y \times N_{A+B}$.

Finally, there are a few other components of the U-Net in Figure 8 that are not included in our original toy example: batch normalization (BN) and 1×1 convolutions. BN is a surprisingly powerful operation that is frequently used after every convolutional layer [17]. For each iteration of training (i.e., for each batch of training samples), BN causes the output values of a given layer to have a fixed mean and variance. This means that the activation maps are scaled such that the average value of all the elements in the activation maps is equal to some number (this number is a learned weight and changes with each batch) and the variance of the activation maps is equal to some number (also a learned weight). This prevents the values in the activation maps from drifting to large difficult-to-handle numbers during training. Empirically, BN results in faster convergence and helps regularize a network. The 1×1 convolution, which is applied in the last layer of the U-Net in Figure 8, is a means of converting a stack of activation maps into a single 2D output image: $N_x \times N_y \times N_z \rightarrow N_x \times N_y$. Conceptually, every individual *X-Y* index in the activation maps prior to the last layer will have multiple values in the z-dimension, $N_z$, one for each activation map. A 1×1 convolution performs a weighted sum of those $N_z$ values to produce a single *X-Y* output. The weights of this weighted sum are learned during training. Consequently, the 1×1 convolution is used as a convenient tool for changing the number of channels of a layer.

The U-Net can be used for more than just segmentation. By simply changing the last layer activation function from sigmoid (binary output) to linear (continuously valued output), and using an appropriate loss function and training set, the U-Net can perform image synthesis. BN often needs to be removed for image synthesis applications, however, as BN can make it difficult to produce consistent quantitative outputs due to the constantly shifting values of the activation maps. The U-Net has been used in a number of image synthesis applications, such as CT synthesis for attenuation correction [2].

## VII. OTHER NETWORKS AND MODELS

CNN's, including the U-net, have gained wide popularity in nuclear medicine, but they are not always the best option for every application. Furthermore, not all CNNs are designed to perform the same function. Developers must select a model type and architecture (anatomy) matched with the proper functionality (physiology) that is appropriate for their prediction task. For example, a model designed for classification will often not be suitable for image synthesis.

Table 2 shows several different classes of AI models and their typical uses. It also lists examples of published architectures for each class. The structure and function of each type of model dictates which applications it is suitable for. For example, a radiomics model should be designed to take a list of numerical values (radiomics features) as input and produce a numerical output (e.g., risk score). Hence, radiomics models often use artificial neural networks (ANN) or decision forests with the appropriate structure (e.g., number of input/output nodes) and function (e.g., activation functions) [18].

Novel classes of model are continuing to be developed. For example, a generative adversarial networks (GAN) is a unique type of model that is gaining popularity in medical imaging. GANs can often perform the same tasks as CNNs but may require fewer labeled data samples [19]. GANs pit two networks against one another (often two CNNs). One network is tasked with producing realistic and accurate outputs (e.g., segmentation masks) and the other network is tasked with predicting whether a sample was generated by the first network or came from a set of true labels. Each network gets penalized when the other network succeeds, which causes both networks to compete and improve. Novel unsupervised and semi-supervised model types are also showing promise. Readers are encouraged to explore benchmark datasets and data science competitions to become familiar with the ever-growing variety of AI architectures [20].

## VIII. CONCLUSIONS

As the foothold of AI within PET imaging continues to grow, more and more people in nuclear medicine will be exposed to concepts and principles of AI. Here, we have described the core structure and function of CNNs, starting basic and then building up to the more complicated U-Net. We hope to have established a foundation of knowledge that readers can build upon and

Table 2. Different classes of AI models and their typical uses.

| Model classes | Input | Output | Typical uses | Example architectures |
|---|---|---|---|---|
| Convolutional neural network (CNN) | | | | |
|     Deep CNN | Image | Class, score* | Image classification, risk score prediction | AlexNet, ResNet |
|     Encoder-decoder CNN | Image | Image, mask | Image segmentation, image synthesis | U-Net, SegNet |
|     Detection CNN | Image | Pixel indices | Object detection/localization | R-CNN, YOLO |
| Generative adversarial network (GAN) | Image | Image, mask | Image-to-image translation | CycleGAN, pix2pix |
| Artificial neural network (ANN) | | | | |
|     Feed forward ANN | Features | Class, score | Radiomics, risk score prediction | Perceptron, multi-layer perceptron |
|     Recurrent ANN | Features | Class, score, text | Language modeling, time series | LSTM |
|     Transformer | Features | Class, score, text | Language modeling | BERT, GPT-3 |
| Decision forest (DF) | | | | |
|     Random forests | Features | Class, score | Radiomics, risk score prediction, classification | CART, bagged trees |
|     Gradient-boosted DF | Features | Class, score | Radiomics, risk score prediction, classification | XGBoost |
| Clustering | Features | Class | Unsupervised classification | k-means, k-NN, |
| Support vector machines (SVM) | Features | Class | Classification | RBF SVM |

*score can mean any continuous variable, such as age, size, risk, etc.

familiarized readers with the building blocks that are used in most machine learning applications. With a greater familiarity of AI principles, readers will be better positioned to develop, evaluate, and use AI tools in future clinical practice.